\documentclass[11pt]{article}

\usepackage[final]{acl}
\usepackage{float} 

\usepackage{times}
\usepackage{latexsym}
\usepackage{booktabs}
\usepackage{amsmath}
\usepackage{multirow}
\usepackage{subcaption}
\usepackage{tabularx}

\usepackage{tikz}
\usetikzlibrary{arrows.meta, positioning, fit}

\usepackage[T1]{fontenc}
\usepackage[utf8]{inputenc}

\usepackage{microtype}

\usepackage{inconsolata}

\usepackage{graphicx}

\title{V-FiLLM: Verified Financial LLM Reasoning Benchmark}

\author{
  Alicia Larsen$^{1*}$ \quad
  Victoire Laurent$^{1*}$ \quad
  Aulia Kharis Rakhmasari$^{1*}$ \\
  \textbf{Lara Turgut}$^{1*}$ \quad
  \textbf{Nino Antulov-Fantulin}$^{1,2}$ \\
  $^{1}$ETH Zürich, Zürich \\
  $^{2}$Aisot Technologies Ltd \\
  \texttt{\{alarsen, vlaurent, arakhmasari, lturgut\}@ethz.ch} \\
  \texttt{nino@aisot.com}
}
\begin{document}
\maketitle
\begin{abstract}
While existing benchmarks have made substantial progress in evaluating LLMs across STEM domains, financial reasoning over structured data remains comparatively less explored. 
We introduce \textbf{V-FiLLM}, a framework that generates financial reasoning benchmarks from executable computation trees grounded in real tables, yielding items whose answers are correct by construction. Trees are evaluated symbolically to obtain ground truth and rendered into natural-language questions, removing any model from the labeling loop, so items can be generated at arbitrary scale without annotation cost and without inheriting a generator's error rate. V-FiLLM exposes four independently controllable axes of difficulty including computation depth, expression breadth, financial concept complexity, and context size. By evaluating on open-source models, we find that accuracy  falls up to 51\% as reasoning depth increases, and up to 47\% points under adversarial numerical perturbations, highlighting remaining challenges in robust financial reasoning over tables. We further show that lightweight LoRA fine-tuning on verified chain-of-thought traces improves accuracy from 81.1\% to 85.6\% on held-out problems and outperforms the base model by 5\% points on FinQA \cite{chen_finqa_2022}, suggesting that targeted, low-cost adaptation is a promising direction for compositional reasoning in financial QA. \footnote{Code available at \url{https://github.com/auliakharis/ML-in-Finance-and-Complex-System}}
\end{abstract}

\section{Introduction}

Recent advances in graph learning~\cite{antulov-fantulin_graph_2026}, transformer language models~\cite{jazbec_impact_2021}, and generally deep learning~\cite{antulov-fantulin_advances_2023} have had a profound influence on financial decision making. However, reasoning over structured financial data remains a distinctive challenge for large language models. Unlike open-ended text generation or even mathematical problem-solving over symbolic expressions, financial reasoning demands the simultaneous satisfaction of several constraints: correct retrieval of values from structured tables, knowledge of domain-specific financial concepts and their formulas, and the ability to compose multi-step arithmetic across heterogeneous quantities such as monetary amounts, rates, and ratios.

Existing benchmarks have made substantial progress in evaluating LLMs on STEM
reasoning and general numerical tasks~\cite{yuan_gsm8k-v_2025,li_cmmlu_2024,rein_gpqa_2023}, yet financial
reasoning over structured tabular data ( particularly in settings that require
compositional depth and domain knowledge) remains comparatively
underexplored~\cite{chen_finqa_2022,zhu_tat-qa_2021,zhao_multihiertt_2022,koncel-kedziorski_bizbench_2024}. A core obstacle to
progress in this area is the lack of benchmarks with explicit, controllable
difficulty~\cite{zhu_dyval_2024,mirzadeh_gsm-symbolic_2025}. Most existing financial QA datasets are
constructed from real-world documents via human annotation or model-assisted
extraction, which conflates
source-level noise (including formatting irregularities, missing values, and
ambiguous phrasing) with genuine failures in financial
reasoning~\cite{chen_hybridqa_2020}. This makes it difficult to isolate where and why
models fail~\cite{dziri_faith_2023}. Furthermore, because ground-truth answers are
written manually or extracted heuristically, the cost of scaling such benchmarks
to cover a broad distribution of reasoning depths is prohibitive~\cite{chen_tabfact_2020}.

We address these limitations by introducing a deterministic benchmark generation framework for financial reasoning over synthetic spreadsheets. The framework generates natural-language questions by sampling executable computation trees grounded in spreadsheet values, producing automatically verified answers without manual annotation. By controlling reasoning depth and incorporating derived financial concepts such as gross profit, operating income, and net income, it enables scalable evaluation of both compositional reasoning and financial domain knowledge.

Using this framework, we construct single-turn and multi-turn benchmarks and evaluate a range of open-source LLMs across increasing reasoning complexity. We show that performance declines sharply with reasoning depth, that LoRA fine-tuning on verified chain-of-thought traces improves Qwen3-4B over its zero-shot baseline, and that OCR-style character corruption is the most effective adversarial perturbation. Together, these results provide a reproducible benchmark for studying robust compositional financial reasoning.

\section{Related Work}

Financial question answering benchmarks have progressively increased in complexity, from multi-step numerical reasoning over reports in FinQA \cite{chen_finqa_2022}, to conversational settings in ConvFinQA \cite{chen_convfinqa_2022}, to hybrid table-text evidence in TAT-QA \cite{zhu_tat-qa_2021}, and long-context retrieval over SEC filings in FinanceBench \cite{islam_financebench_2023}. FINGPT Bench \cite{wang_fingpt_2023} offers broader financial NLP evaluation across sentiment, relation, and forecasting tasks. These datasets are constructed via human annotation, which limits scalability and makes reasoning difficulty difficult to control systematically.

General reasoning benchmarks such as DROP \cite{dua_drop_2019} and GSM8K \cite{yuan_gsm8k-v_2025} have been widely used to probe arithmetic and multi-step mathematical reasoning, while TabFact \cite{chen_tabfact_2020} addresses fact verification over semi-structured tables. Neither line of work targets the intersection of domain-specific financial knowledge and compositional tabular reasoning that motivates our benchmark.

Our framework addresses these gaps through a deterministic, annotation-free generation pipeline with explicit control over reasoning depth, grounded in synthetic financial spreadsheets and verified computation trees.

\section{Benchmark Generation Pipeline}

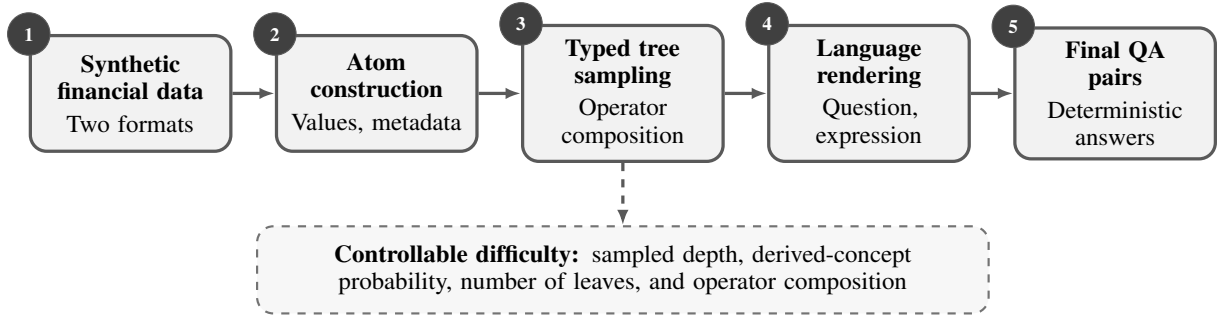
\begin{figure*}[t]
\centering
\resizebox{\textwidth}{!}{%
\begin{tikzpicture}[
    font=\footnotesize,
    node distance=0.55cm,
    stage/.style={
        rectangle, rounded corners=6pt,
        draw=black!65, very thick, align=center,
        text width=2.25cm, minimum height=1.5cm,
        fill=black!5, inner sep=5pt
    },
    badge/.style={
        circle, draw=black!65, fill=black!75, text=white,
        font=\bfseries\scriptsize,
        minimum size=5.5mm, inner sep=0pt
    },
    arrow/.style={-{Latex[length=2mm]}, very thick, draw=black!65},
    difficulty/.style={
        rectangle, rounded corners=6pt,
        draw=black!55, dashed, thick, align=center,
        text width=9cm, fill=black!3, inner sep=7pt
    }
]
\node[stage] (data) {\textbf{Synthetic}\\\textbf{financial data}\\[2pt]
    Two formats};
\node[stage, right=of data] (atoms) {\textbf{Atom}\\\textbf{construction}\\[2pt]
    Values, metadata};
\node[stage, right=of atoms] (tree) {\textbf{Typed tree}\\\textbf{sampling}\\[2pt]
    Operator composition};
\node[stage, right=of tree] (language) {\textbf{Language}\\\textbf{rendering}\\[2pt]
    Question, expression};
\node[stage, right=of language] (qa) {\textbf{Final QA}\\\textbf{pairs}\\[2pt]
    Deterministic answers};

\foreach \n/\i in {data/1, atoms/2, tree/3, language/4, qa/5}
    \node[badge, above left=-2mm and -2mm of \n] {\i};

\foreach \a/\b in {data/atoms, atoms/tree, tree/language, language/qa}
    \draw[arrow] (\a) -- (\b);

\node[fit=(data)(atoms)(tree)(language)(qa), inner sep=0pt] (pipelinebox) {};
\node[difficulty, below=0.8cm of pipelinebox.south] (control) {%
    \textbf{Controllable difficulty:} sampled depth,
    derived-concept probability, number of leaves,
    and operator composition};
\draw[arrow, dashed] (tree.south) -- (control.north -| tree.south);
\end{tikzpicture}%
}
\caption{Benchmark generation pipeline. Synthetic spreadsheets
are decomposed into typed atoms, composed into expression trees
whose depth and operator mix control problem difficulty, and
rendered into natural-language questions paired with
deterministically computed answers.}
\label{fig:benchmark-pipeline}
\end{figure*}

We generate the benchmark using a deterministic pipeline that maps synthetic financial spreadsheets to natural-language questions with executable answers. Starting from structured financial data, the pipeline builds typed semantic representations of spreadsheet cells, samples computation graphs with controlled depth and breadth, grounds each graph in concrete values, and renders the result as a financial reasoning question. This ensures that each example has a known computation trace and an automatically verifiable answer. 

\subsection{Financial Data}
The first stage constructs two synthetic financial data sources. The first mimics a typical 10-Q filing, with financial tables resembling those found in company reports. The second is a regularized financial sheet with consistent column names, units, and temporal structure, allowing us to isolate compositional numerical reasoning from formatting noise, extraction artifacts, and missing values.

The regularized sheet contains company-year observations for 15 synthetic companies from 2020 to 2025. Each row includes company metadata, such as name, ticker, sector, country, exchange, and credit rating, followed by normalized financial variables including revenue, costs, expenses, assets, liabilities, equity, cash, receivables, inventories, investments, current liabilities, capital expenditures, dividends, shares outstanding, stock price, and employees. The number of companies is adjustable; we use 6 in the prompt to balance realistic report scale with manageable LLM context length.

Financial values are generated to preserve internal coherence while remaining synthetic. Revenue is initialized from company-specific ranges and evolves through sampled yearly growth or decline. Income-statement quantities are tied to revenue through plausible ratios, while balance-sheet components are sampled as percentages of related parent quantities, such as cash relative to total assets or inventories relative to cost of goods sold. Additional variables are sampled from realistic ranges or derived from intermediate quantities, yielding tables that support meaningful arithmetic and compositional reasoning.

 From these spreadsheets, we construct typed financial atoms. Each atom corresponds to one spreadsheet cell and stores its numerical value together with semantic metadata: financial concept, company, fiscal year, unit, and quantity type. These atoms provide the grounding layer for question generation.
 
\subsection{Question-Answer Pair Generation}

The next stage samples a symbolic expression template, represented as a typed binary tree. Leaves are financial atoms, while internal nodes specify an operation and metadata about their children. The operator set includes addition, subtraction, multiplication, ratios, growth rates, minimum, maximum, and average. The sampler is type-aware: it only constructs expressions whose child outputs are compatible with the parent operation and each other, preventing invalid expressions such as multiplying a year by a monetary amount.

\begin{figure}[ht]
\centering
\resizebox{1.05\linewidth}{!}{%
\begin{tikzpicture}[
    font=\small,
    op/.style={
        circle,
        draw=black!70,
        very thick,
        fill=blue!8,
        minimum size=9mm,
        align=center
    },
    leaf/.style={
        rectangle,
        rounded corners=4pt,
        draw=black!60,
        thick,
        fill=gray!8,
        minimum width=2.9cm,
        minimum height=8mm,
        align=center
    },
    edge/.style={
        -{Latex[length=2mm]},
        thick,
        draw=black!60
    },
    title/.style={
        font=\bfseries,
        align=center
    },
    expr/.style={
        align=center,
        font=\footnotesize,
        text width=6.4cm
    }
]

% =========================================================
% Example 0: Depth-0 expression
% =========================================================

\node[title] (t0title) at (0, 0) {Depth 0};

\node[leaf, below=0.5cm of t0title] (t0leaf)
{Revenue\\
\scriptsize Apex Dynamics Corp\\
\scriptsize 2020};

\node[expr, below=0.35cm of t0leaf, xshift=-0.3cm] (t0expr) {
\emph{What is revenue for Apex Dynamics Corp in 2020?}
};

% =========================================================
% Example 1: Depth-1 expression
% Shifted lower
% =========================================================

\begin{scope}[shift={(0,-3.7)}]

\node[title] (t1title) at (0, 0) {Depth 1};

\node[op, below=0.5cm of t1title] (t1root) {$\times$};

\node[leaf, below left=0.45cm and 0.45cm of t1root] (t1a)
{Current liabilities\\
\scriptsize Crimson Pharmaceuticals\\
\scriptsize 2021};

\node[leaf, below right=0.45cm and 0.45cm of t1root] (t1b)
{Income tax rate\\
\scriptsize Crimson Pharmaceuticals\\
\scriptsize 2021};

\draw[edge] (t1root) -- (t1a);
\draw[edge] (t1root) -- (t1b);

\node[expr, below=0.55cm of t1root, yshift=-1.5cm, xshift=-0.3cm] (t1expr) {
\emph{What is current liabilities scaled by income tax for Crimson Pharmaceuticals in 2021?}
};

\end{scope}

% =========================================================
% Example 2: Depth-2 expression
% Kept the same
% =========================================================

\node[title] (t2title) at (7.8, 0) {Depth 2};

\node[op, below=0.5cm of t2title] (t2root) {$-$};

\node[op, below left=0.3cm and 0.8cm of t2root] (t2left) {$\times$};
\node[op, below right=1.6cm and 1.8cm of t2root] (t2right) {$\times$};

\node[leaf, below=0.85cm of t2left, xshift=-2.0cm] (t2a)
{Total assets\\
\scriptsize Meridian Systems \\
\scriptsize 2023};

\node[leaf, below=0.85cm of t2left, xshift=1.0cm] (t2b)
{Income tax rate\\
\scriptsize Meridian Systems \\
\scriptsize 2023};

\node[leaf, below=0.95cm of t2right, xshift=-2.0cm] (t2c)
{COGS\\
\scriptsize Meridian Systems \\
\scriptsize 2023};

\node[leaf, below=0.95cm of t2right, xshift=1.0cm] (t2d)
{Income tax rate\\
\scriptsize Meridian Systems \\
\scriptsize 2023};

\draw[edge] (t2root) -- (t2left);
\draw[edge] (t2root) -- (t2right);

\draw[edge] (t2left) -- (t2a);
\draw[edge] (t2left) -- (t2b);

\draw[edge] (t2right) -- (t2c);
\draw[edge] (t2right) -- (t2d);

\node[expr, below=1.55cm of t2root, yshift=-3.25cm] (t2expr) {
\emph{What is the difference between total assets scaled by income tax and cost of goods sold scaled by income tax for Meridian Systems in 2023?}
};

\end{tikzpicture}%
}
\caption{Examples of grounded computation trees and their rendered natural-language questions. Leaves include spreadsheet metadata such as concept, company, and year; internal nodes define the arithmetic computation.}
\label{fig:grounded-tree-examples}
\end{figure}
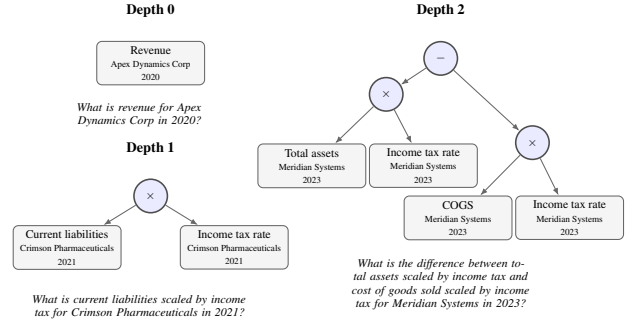

A key feature of this generation process is that because of the tree structure, the difficulty of a question can be explicitly controlled. In particular, tree depth provides a direct proxy for the number of reasoning steps required.  Indeed, simple questions such as: ''\textit{Provide value of company A’s total assets in 2024}" are represented as a simple leaf and have depth 0, while slightly more complex ones such as: ''\textit{Give Maximum between company A’s total assets and company B’s total assets in 2024}" are represented as a tree with two leaves and one operator node and thus have depth 1.  Shallow expressions correspond to simple lookups or one-step arithmetic questions, while deeper expressions require the model to compose several intermediate results. The pipeline can also optionally use balanced trees, which distribute computation more evenly across branches. This allows us to also control the breadth of a question, as opposed to the depth, with a breadth to depth growth factor of $2^i$ for depth $i$. See \ref{fig:grounded-tree-examples} for an example of such trees.

The pipeline also supports named derived financial concepts, such as gross profit, operating income, pretax income, income tax expense, net income, current assets, long-term assets, and long-term liabilities. These concepts are stored as special atoms with hidden executable formula trees. Because their formulas are not shown in the question and the values cannot be directly looked up, they test financial knowledge in addition to arithmetic reasoning.

To avoid ambiguity between named financial concepts and arbitrary equivalent arithmetic expressions, the sampler applies a rejection step for protected derived concepts. If a sampled template reconstructs the canonical formula of a protected concept without explicitly using the corresponding named concept atom, the tree is rejected and resampled. This prevents, for example, a manually sampled difference between revenue and cost of goods sold from being treated differently than the named concept ''gross profit". As a result, the benchmark maintains a clear distinction between raw arithmetic composition and explicitly introduced financial abstractions.

The sampling process is parameterized to control the distribution of generated examples. In particular, we specify a minimum and maximum template depth, and each question is sampled with a depth drawn from this range. This makes it possible to generate mixtures of simple and highly compositional questions, or to focus evaluation on a specific difficulty band. We also control the probability of sampling named derived concepts, in order to ensure that they appear more often than regular spreadsheet cells.

After a valid template is sampled, the pipeline grounds it in concrete spreadsheet atoms using a binding environment that tracks constraints such as company, fiscal year, and financial concept. Same-period operations bind operands to compatible company-year contexts. Growth operations select two distinct years for the same company and concept, while temporal aggregations such as minimum, maximum, and average select a set of reporting years for a given company-concept pair. The result is a fully bound expression whose leaves are spreadsheet cells and whose internal nodes define the exact computation.

The bound expression is then rendered into a natural-language question through a bottom-up semantic procedure. Leaves are translated into short phrases using their metadata, such as ''\textit{revenue for Apex Dynamics Corp in 2020}”. Internal nodes then combine their children with operator-specific templates; for example, a subtraction node over revenue and cost of goods sold becomes “\textit{the difference between revenue and cost of goods sold for Apex Dynamics Corp in 2020}”. Because intermediate phrases retain metadata such as company, year, unit, and concept, the renderer can avoid redundant phrasing when context is shared across subexpressions.

After rendering, we apply lightweight linguistic data augmentation to the natural-language questions. 
These augmentations preserve the underlying computation graph and ground-truth answer, but introduce surface-level variation in how the question is phrased. For example, we introduce minor typographical errors, vary capitalization, and replace standard phrasing with informal alternatives such as ''\textit{What is the value of ...}” versus ''\textit{How would you calculate ...}”. Such perturbations enable the assessment of model robustness to superficial linguistic variations and reduce the likelihood that benchmark performance is driven by sensitivity to specific prompt templates. Evaluating model behavior across multiple semantically equivalent formulations provides a more reliable measure of financial reasoning and domain knowledge, while improving the ecological validity of the benchmark by more closely reflecting the diversity of expressions observed in practical financial applications.

The executable expression linked to the rendered question is evaluated directly to obtain the ground-truth answer. Since the answer is computed from the expression tree rather than written manually, each question-answer pair is automatically verified. For every generated example, we store the natural-language question, the numeric answer, the full symbolic expression, the template expression, the requested and realized depths, the probability of sampling derived concepts, and the list of spreadsheet cells used as leaves. This metadata makes the benchmark reproducible and enables fine-grained analysis by depth, operator type, number of required cells, use of derived concepts, and temporal reasoning structure.

\subsection{Augmentation}
When constructing a financial benchmark for large language models (LLMs), it is
important to account for the linguistic variability that naturally arises in
real-world user interactions. Semantically equivalent financial questions can be
expressed through a variety of surface-level formulations, including differences
in phrasing, e.g., ``What is the value of...'' versus ``How would you
calculate...'', grammar, capitalization, punctuation, and formatting. Prior work
has shown that LLMs are highly sensitive to such meaning-preserving variation:
\citet{sclar_quantifying_2023} report performance swings of up to 76 accuracy points from
prompt formatting alone, and \citet{mizrahi_state_2024} find that instruction
paraphrases change both absolute and relative model rankings. Controlled input
augmentations were therefore applied to preserve the underlying financial task
while altering its textual representation, following the invariance-testing
paradigm of \citet{ribeiro_beyond_2020}. Such augmentations enable the assessment of model
robustness to superficial linguistic variations and reduce the likelihood that
benchmark performance is driven by sensitivity to specific prompt templates.
Evaluating model behavior across multiple semantically equivalent formulations
provides a more reliable measure of financial reasoning and domain knowledge,
while improving the ecological validity of the benchmark by more closely
reflecting the diversity of expressions observed in practical financial
applications.

\section{Extensions}

% i tried moving it into the pipeline genration area like we said, but left it here incase we don't like it and want it back like this

\subsection{Multi-Turn Extension}
\label{sec:multiturn}

Single-turn financial QA conflates two distinct failure modes: errors in
intermediate arithmetic and errors in compositional planning. Prior work on
compositional reasoning finds that transformers degrade sharply as problem depth
grows, in ways that aggregate accuracy alone does not expose
\citep{dziri_faith_2023}, and decomposition-based prompting improves performance
precisely by making the sub-problem sequence explicit \citep{zhou_least--most_2023}. Yet
model performance also drops substantially when the same task is distributed
across conversational turns rather than posed at once \citep{laban_llms_2025},
making the multi-turn setting a distinct axis of evaluation rather than a
reformatting of the single-turn one.

Let each problem be associated with an expression tree $T$ of bound depth
$d(T)$, where internal nodes correspond to parenthesized sub-expressions in the
string representation form of $T$. We convert $T$ into a multi-turn dialogue under the
following rules, we convert only trees with $d(T) \geq 1$, retaining shallower items unchanged.
Each parenthesized sub-expression becomes one turn, ordered by computational
dependency; the final turn restates the original question in full, keeping the
multi-turn variant comparable to its one-shot counterpart. Per-turn targets are
read directly from the corresponding nodes of $T$, requiring no additional
annotation.

Because the procedure is purely structural, the multi-turn benchmark inherits
the scale and coverage of the single-turn benchmark while exposing
intermediate computations as first-class evaluation targets. This enables per-depth diagnostics
that are not identifiable from single-question correctness.

\begin{table}[t]
\centering
\small
\begin{tabular}{@{}p{\columnwidth}@{}}
\toprule
\textbf{Original question (single-turn).} What is the
operating income for NovaStar Technologies in 2024? \\[2pt]
\textbf{Expression tree.}
\texttt{subtract(subtract(revenue, cogs), operating\_expenses)}
\quad ($d(T)=2$) \\
\midrule
\textbf{Turn 1 --- User:} What is the gross profit for
NovaStar Technologies in 2024? \\
\textbf{Turn 1 --- Model:} \$100,535 \\[2pt]
\textbf{Turn 2 --- User:} What is the operating income for
NovaStar Technologies in 2024? \\
\textbf{Turn 2 --- Model:} \$80,626 \\
\bottomrule
\end{tabular}
\caption{Multi-turn conversion of a depth-2 problem. Each
turn resolves one node of the expression tree, with the final
turn restating the original single-turn question.}
\label{tab:multiturn-example}
\end{table}

\subsection{Adversarial Robustness}
\label{sec:adversarial}

A deployed financial QA system encounters tables that are noisy, partially
missing, or assembled from heterogeneous sources. Following prior work on
adversarial robustness in table QA \citep{zhao_robut_2023,zhou_freb-tqa_2024}, we perturb the input
tables of existing items while leaving the question and answer unchanged. \citet{mirzadeh_gsm-symbolic_2025} report drops of up to 65\% from a single such clause. All
perturbations are restricted to \emph{question-irrelevant} cells, i.e., cells
absent from the expression tree computing the answer, so the ground truth is
invariant by construction and any accuracy drop reflects model failure rather
than label shift.

Four perturbation families mirror failure modes in real financial pipelines.

\textbf{(i) Missing values:} random question-irrelevant cells are emptied,
simulating incomplete filings and partial extraction. 

\textbf{(ii) Garbage
values:} such cells are overwritten with $\{-10^7,\ 10^{12},\ -1,\ \texttt{"ERROR"}\}$, probing whether the model anchors to relevant cells or is
drawn to salient outliers. 

\textbf{(iii) OCR look-alikes:} digits are replaced
with visually similar characters (\texttt{0}$\to$\texttt{O}, \texttt{1}$\to$\texttt{l},
\texttt{5}$\to$\texttt{S}, etc.), reproducing a common OCR failure on scanned
statements.   

\textbf{(iv) Cross-sheet contamination:} rows are swapped with rows
from other companies, yielding tables that are well-formed but semantically
inconsistent, as in faulty multi-document consolidation.

We additionally perturb at the question level, appending irrelevant context
(rumors, market sentiment, statements about other companies). This targets the
distractibility documented by \citet{shi_large_2023}, who find that a single irrelevant
sentence substantially degrades arithmetic reasoning.

\subsection{Controlling Reasoning Difficulty}
\label{sec:magnitude_depth}
Because each question is derived from an expression tree over table cells, we can
vary computational structure and numerical properties independently while
retaining exact ground truth.

\textbf{(i) Depth.} Tree depth sets the number of compositional steps, increasing
demands on multi-step reasoning and intermediate-state tracking.

\textbf{(ii) Breadth.} Balanced trees, rather than linear chains, force several
sub-results to be computed and held simultaneously, widening the set of relevant
table entries.

\textbf{(iii) Value scaling.} A multiplicative factor enlarges operands without
altering reasoning structure, varying numerical difficulty independently of depth.

\subsection{LoRA}
To improve the model's ability to perform the domain-specific numerical reasoning and calculation tasks required by our financial benchmark, a pretrained model can be fine-tuned using Low-Rank Adaptation (LoRA) \cite{hu_lora_2021}. Rather than updating
the full set of model parameters, LoRA introduces trainable low-rank matrices
while keeping the original model weights frozen, following a broader line of
parameter-efficient fine-tuning methods \citep{houlsby_parameter-efficient_2019, lialin_scaling_2024}. In our implementation, LoRA was applied to the attention projection layers (\verb|q_proj|, \verb|k_proj|, \verb|v_proj|, and \verb|o_proj|) as well as the feed-forward network projections (\verb|gate_proj|, \verb|up_proj|, and \verb|down_proj|). Adapting the attention layers allows the model to learn task-specific patterns in how financial information is represented and attended to, while modifying the feed-forward projections enables adjustments to the model’s internal feature transformations and reasoning capabilities. A rank of ($r=16$), a scaling factor ($\alpha=32$), and a dropout rate of $0.05$ were used, to provide a balance between adaptation capacity and parameter efficiency. This configuration substantially reduced the number of trainable parameters and the associated computational requirements while enabling effective specialization of the base model for financial reasoning tasks.

\section{Results}
\subsection{Benchmarking results}

\paragraph{Model comparison} Table~\ref{tab:main} reports accuracy for six models on 250 mixed-depth questions, over real 10-Q filings and over simplified financial statements. The 10-Q Fillings type support reasoning trees only up to depth 4, whereas simplified statements admit depths up to 8. In the mixed-depth setting, we sample questions with depths between zero and five for the simplified case and zero and four for 10Q.

Model rankings remain stable across both evaluation settings. Gemma-31B achieves the highest accuracy in both contexts (98.4\% and 97.6\%), closely followed by GPT-OSS-120B, Qwen3.7-Plus, and DeepSeek-v4-Flash. The performance gap widens significantly on simplified statements that introduce more complex queries. In this setting, Llama-3.3-70B and Qwen3.5-9B degrade to 80.8\% and 55.2\% respectively, whereas Gemma-31B and DeepSeek-v4-Flash exhibit high resilience, dropping less than a percentage point compared to the filings baseline.

Despite a significantly smaller parameter count, Gemma-31B matches or outperforms both GPT-OSS-120B and Qwen3.7-Plus across all conditions. This indicates that task effectiveness is likely driven more by specialized training data and robust table parsing mechanisms than by model scale. 

\paragraph{Multi-turn Results} Restructuring complex prompts into sequential sub-queries mitigates much of the performance degradation observed on the harder dataset. Under this multi-turn framework, Qwen3.5-9B improves from 55.2\% to 86.8\% and Llama-3.3-70B rises from 80.8\% to 86.8\%, whereas models already operating near the performance ceiling exhibit negligible shifts. The greatest gains manifest in models that struggled most in the single-turn setting. We hypothesize that query decomposition ease the context load required for single-step reasoning.

\paragraph{Effect of reasoning depth.} Table~\ref{tab:depth} shows the accuracy based on the number of steps needed to find information and do the calculation. We tested 100 questions for each level.  On the simplified statements, accuracy stays steady up to 6 steps, but then falls quickly. For example, Gemma-31B drops from 85.0\% at 6 steps to 55.0\% at 8 steps, and DeepSeek-v4-Flash drops from 84.0\% to 26.0\%. DeepSeek-v4-Flash gets worse much faster than Gemma-31B after 6 steps, which flips their ranking from the easier questions.

For the 10-Q fillings, where only depths 1--4 are available, Both models get almost perfect scores for 1 or 2 steps, but drop to about 60--72\% for 3 or 4 steps. When comparing the same number of steps, the 10-Q fillings are about 20 points harder than the simplified ones at steps 3 and 4. This is expected because real documents have messy layouts, footnotes, and mixed units that the simplified versions do not have.

% ---------------------------------------------------------------------
% TABLE 1 — Main results
% ---------------------------------------------------------------------
\begin{table}[!h]
\centering
\small
\setlength{\tabcolsep}{6pt}
\begin{tabularx}{\columnwidth}{@{}Xrrr@{}}
%\begin{tabular}{@{}lrrr@{}}
\toprule
\multirow{2}{*}{\textbf{Model}} & \multicolumn{2}{c}{\textbf{Single-turn}} & \textbf{Multi-turn} \\
\cmidrule(lr){2-3} \cmidrule(lr){4-4}
 & 10Q & Simplified & Simplified \\
\midrule
\multicolumn{4}{@{}l}{\textit{Large ($\geq$70B)}} \\
GPT-OSS-120B        & 98.0 & 93.6 & \textbf{97.6} \\
Qwen3.7-Plus        & 98.0 & 94.4 & \textbf{98.0} \\
DeepSeek-v4-Flash   & 96.4 & 95.6 & 88.0 \\
Llama-3.3-70B       & 97.2 & 80.8 & 86.8 \\
\midrule
\multicolumn{4}{@{}l}{\textit{Small / mid ($\leq$31B)}} \\
Gemma-31B           & \textbf{98.4} & \textbf{97.6} & 96.4 \\
Qwen3.5-9B          & 90.4 & 55.2 & 86.8 \\
\bottomrule
%\end{tabular}
\end{tabularx}
\caption{Accuracy (\%) on 250 mixed-depth questions under 10Q-fillings and simplified spreadsheet context.
Multi-turn decomposes each question into sequential sub-queries.
Best per column in \textbf{bold}.}
\label{tab:main}
\end{table}

% ---------------------------------------------------------------------
% TABLE 2 — Reasoning depth
% ---------------------------------------------------------------------
\begin{table}[!h]
\centering
\small
\setlength{\tabcolsep}{4pt}
\begin{tabularx}{\columnwidth}{@{}l*{4}{>{\centering\arraybackslash}X}@{}}
%\begin{tabularx}{\columnwidth}{@{}lrrrr@{}}
% \begin{tabular}{@{}lrrrr@{}}
\toprule
\multirow{2}{*}{\textbf{Depth}} & \multicolumn{2}{c}{\textbf{10Q}} & \multicolumn{2}{c}{\textbf{Simplified}} \\
\cmidrule(lr){2-3} \cmidrule(lr){4-5}
 & Gemma & DS-v4 & Gemma & DS-v4 \\
\midrule
1 & \textbf{100.0} & \textbf{98.0} & 84.0 & 77.0 \\
2 & 98.0 & 95.0 & \textbf{95.3} & 78.0 \\
3 & 70.0 & 60.0 & 90.7 & 86.0 \\
4 & 72.0 & 70.0 & 94.3 & \textbf{93.0} \\
5 & \multicolumn{1}{c}{--} & \multicolumn{1}{c}{--} & 89.7 & 92.0 \\
6 & \multicolumn{1}{c}{--} & \multicolumn{1}{c}{--} & 85.0 & 84.0 \\
7 & \multicolumn{1}{c}{--} & \multicolumn{1}{c}{--} & 72.0 & 53.0 \\
8 & \multicolumn{1}{c}{--} & \multicolumn{1}{c}{--} & 55.0 & 26.0 \\
\midrule
Avg. & 85.0 & 80.8 & 83.3 & 73.6 \\
\bottomrule
% \end{tabular}
\end{tabularx}
\caption{Accuracy (\%) by reasoning depth (number of chained retrieval/arithmetic steps), 100 questions per cell.
Depths 5--8 were not instantiable in the 10Q setting.
DS-v4 = DeepSeek-v4-Flash.}
\label{tab:depth}
\end{table}

\subsection{Adversarial Robustness Results}
\label{sec:adversarial-results}

% ---------------------------------------------------------------------
% TABLE 3 — Adversarial perturbations
% ---------------------------------------------------------------------
\begin{table}[t]
\centering
\small
\setlength{\tabcolsep}{3.5pt}
\begin{tabular}{@{}lrrrr@{}}
\toprule
\multirow{2}{*}{\textbf{Perturbation}} & \multicolumn{2}{c}{\textbf{10Q}} & \multicolumn{2}{c}{\textbf{Simplified}} \\
\cmidrule(lr){2-3} \cmidrule(lr){4-5}
 & Gemma & DS-v4 & Gemma & DS-v4 \\
\midrule
Clean (no perturbation)   & 98.4 & 96.4 & 97.6 & 95.6 \\
\midrule
Missing values            & 78.0 & 72.0 & 59.0 & 44.0 \\
Garbage values            & 82.0 & 70.0 & 54.0 & 42.0 \\
OCR look-alike            & 82.0 & 76.0 & 72.0 & 53.0 \\
Cross-sheet contamination & 85.0 & 78.0 & 58.0 & 44.0 \\
Distractor information    & 82.0 & 79.0 & 77.0 & 65.0 \\
Unit / scale shift        & \phantom{0}3.0 & 22.0 & 57.0 & 38.0 \\
\midrule
Combined                  & 26.0 & 17.0 & 48.0 & 43.0 \\
\midrule
Avg. $\Delta$ vs.\ clean  & $-$27.3 & $-$29.6 & $-$35.7 & $-$47.0 \\
\bottomrule
\end{tabular}
\caption{Accuracy (\%) under adversarial spreadsheet perturbations, mixed depth, 100 questions per cell.
Clean rows are the corresponding mixed-depth baselines from Table~\ref{tab:main}.}
\label{tab:adversarial}
\end{table}

All models perform much worse when we do the pertubations test (Table~\ref{tab:adversarial}). On average, across the six different types of noise, Gemma-31B's score drops by 27.3 points on 10-Q Fillings and DeepSeek-v4-Flash drops by 29.6 points. These drops are even bigger on the simplified statements (35.7 and 47.0 points). The ranking of the models always stays the same: Gemma-31B beats DeepSeek-v4-Flash on all seven noise tests in both document types, and its lead actually grows when the test gets harder.

Adding useless extra information is the easiest test. It lowers both models' scores by less than 20 points on 10-Q Fillings, showing that they are pretty good at ignoring extra text that does not matter. Other issues, like missing numbers, random text, scanning typos, and mixed-up spreadsheets, all result in scores between 70\% and 85\% on 10-Q Fillings. This means that mild noise and confusing data lower the scores but do not completely break the models.

Two situations are completely different. Changing the units or scale is the most harmful test on 10-Q Fillings, crashing Gemma-31B to just 3.0\% and DeepSeek-v4-Flash to 22.0\%. This massive failure shows that both models mostly read numbers without checking the unit labels at the top of the table. Interestingly, changing the units is much less harmful on the simplified statements (where they score 57.0\% and 38.0\%). This again demonstrates that the real documents are harder than the simplified ones, likely because 10-Q Fillings have much more complicated and messy unit labels. 

Finally, when we combine all the noise together, the scores fall to 26.0\% and 17.0\% on real documents. This is worse than almost any single problem on its own (except the unit changes), meaning the different errors pile up and make things much harder. In this combined test, the gap between the two models shrinks to just 9 points, suggesting the test becomes so messy that even the better model's table-reading skills cannot save it.

\subsection{LoRA Fine-Tuning with Verified Chain-of-Thought}
\label{sec:lora-cot}

\begin{table}[h]
\centering
\small
\begin{tabularx}{\columnwidth}{@{}*{5}{>{\centering\arraybackslash}X}@{}}
%\begin{tabular}{@{}rrrrr@{}}
\toprule
$r$ & $\alpha$ & Dropout & Acc. (\%) & Correct \\
\midrule
\multicolumn{3}{@{}l}{Baseline (no LoRA)} & 81.1 & 73/90 \\
\midrule
8  & 8  & 0.10 & \textbf{85.6} & 77/90 \\
8  & 8  & 0.15 & \textbf{85.6} & 77/90 \\
8  & 4  & 0.10 & 83.3 & 75/90 \\
16 & 32 & 0.05 & 83.3 & 75/90 \\
4  & 4  & 0.10 & 82.2 & 74/90 \\
8  & 16 & 0.10 & 81.1 & 73/90 \\
8  & 16 & 0.15 & 81.1 & 73/90 \\
8  & 4  & 0.15 & 81.1 & 73/90 \\
\bottomrule
%\end{tabular}
\end{tabularx}
\caption{LoRA hyperparameter sweep on the held-out benchmark
($n=90$). Baseline is the zero-shot base model without LoRA
adapters. Configurations are ordered by accuracy; best result
in \textbf{bold}. We vary rank $r \in \{4, 8, 16\}$, scaling
factor $\alpha \in \{4, 8, 16, 32\}$, and dropout
$\in \{0.05, 0.10, 0.15\}$.}
\label{tab:lora-sweep}
\end{table}

\begin{table}[h!]
\centering
\small
\begin{tabularx}{\columnwidth}{@{}*{5}{>{\centering\arraybackslash}X}@{}}
% \begin{tabular}{@{}rrrrr@{}}
\toprule
$r$ & $\alpha$ & Dropout & Acc. (\%) & Correct \\
\midrule
\multicolumn{3}{@{}l}{Baseline (no LoRA)} & 27& 27/100\\
\midrule
8  & 8  & 0.10 & 32& 32/100\\
% \end{tabular}
\end{tabularx}
\caption{Evaluation of LoRA adapters on the FinQA dataset. }
\label{tab:lora-FinQA}
\end{table}

We fine-tune Qwen3.5-4B on the financial QA training pool using a two-phase recipe. \textbf{Phase 1} generates chain-of-thought (CoT) traces for each training item and retains only those whose final answer matches the ground-truth value derived from the expression tree, yielding 608 verified examples from a pool of 688 (88\% retention) \textbf{Phase 2} fine-tunes the model using LoRA adapters (0.50\% trainable parameters) on the verified traces. We sweep $r$, $\alpha$, and dropout on the held-out benchmark ($n{=}90$; Table~\ref{tab:lora-sweep}): $r{=}8$, $\alpha{=}8$, dropout $0.10$ attains the highest accuracy ($85.6\%$), tied at dropout $0.15$. Accuracy is most sensitive to the $\alpha/r$ ratio, with $\alpha/r{\neq}1$ erasing the gain over baseline.

To evaluate generalization to other datasets, we evaluated the LoRA-finetuned Qwen3.5-4B on FinQA \cite{chen_finqa_2022}, another dataset of question-answer pairs based on financial reports. We used the LoRA configuration $r{=}8$, $\alpha{=}8$, and dropout $0.10$, which performed best among the tested settings while using simpler regularization. As shown in Table \ref{tab:lora-FinQA}, the LoRA-finetuned model correctly answered 32/100 questions, which is an increase compared to 27/100 for the baseline. This experiment was conducted solely as a proof of concept to determine whether the LoRA performance improvements stem from overfitting.

\section{Conclusion}

We introduced a deterministic benchmark generation framework for evaluating financial reasoning in large language models over structured tabular data. By generating executable computation trees grounded in synthetic financial spreadsheets, our approach produces automatically verified question-answer pairs while providing explicit control over reasoning depth, expression breadth, financial-domain complexity, and linguistic variation. This enables systematic evaluation of compositional financial reasoning without the scalability and annotation limitations of existing benchmarks.

Our experiments demonstrate that reasoning depth is the primary factor limiting model performance, with accuracy degrading substantially as the number of compositional reasoning steps increases. While LoRA fine-tuning on verified chain-of-thought traces improves performance over strong zero-shot baselines, considerable headroom remains, particularly on deeper reasoning tasks. We further show that OCR-style character corruption poses a greater challenge than missing or irrelevant data, highlighting the importance of robust numerical extraction in practical financial applications. The proposed multi-turn benchmark additionally provides fine-grained insight into intermediate reasoning failures that are not observable through end-to-end evaluation alone.

Overall, the benchmark provides a reproducible and extensively tested metric for studying financial reasoning, model robustness, and reasoning scalability in LLMs. Future work will expand the diversity of financial concepts and document formats, increase coverage of deeper reasoning chains, incorporate cross-table and cross-document reasoning, and explore verifier-guided training and inference methods that further improve reliability on complex financial analysis tasks.
% =====================================================================
%  FinLLM benchmark result tables (ACL / acl_natbib style)
%  Requires in preamble:
%    \usepackage{booktabs}
%    \usepackage{multirow}
%    \usepackage{amssymb}   % for \dag / \ddag footnote marks if needed
%  All tables are single-column (fit the ACL 3.2in column).
% =====================================================================

\section{Limitations}

Although the final answers in our dataset are guaranteed to be correct by construction, the questions themselves may still contain imperfections. In particular, the underlying mathematical reasoning steps may occasionally be translated into English in ways that are ambiguous or lack sufficient context.
Several limitations of our evaluation should also be acknowledged. The main limitation comes from the constrained compute resources for this study.
Our experiments cover only six models, are restricted to English-language documents, and consider a single table-formatting convention. In future work, with access to more compute power, we plan to increase statistical significance by increasing the number of tests and running multiple trials over the same questions.
%In addition, each experiment was conducted only once, meaning that we are unable to report measures of variability or statistical uncertainty.

%Finally, our fine-tuning experiments, which explored hyperparameters such as LoRA rank, scaling factor, and dropout, were conducted using a relatively small evaluation set of 90 problems. Because the best-performing configuration was selected using this same limited set, the resulting performance gains should be interpreted as preliminary evidence that the models can learn effectively from this data, rather than as precise estimates of the magnitude of improvement.
Finally, our fine-tuning experiments—exploring hyperparameters such as LoRA rank, scaling factor, and dropout—were restricted to a relatively small evaluation set of 90 problems due to limited compute resources. Because the best-performing configuration was selected using this same constrained set, the observed performance gains should be interpreted as preliminary evidence that it is possible to improve the reasoning of models, while extensive fine-tuning is left for future work.

\bibliography{references}

\end{document}